\documentclass[10pt,twocolumn,letterpaper]{article}

\usepackage{cvpr}              

\usepackage[dvipsnames]{xcolor}

\definecolor{cvprblue}{rgb}{0.21,0.49,0.74}
\usepackage[pagebackref,breaklinks,colorlinks,citecolor=cvprblue]{hyperref}
\usepackage[]{multirow}
\usepackage{bm}

\def\paperID{10003} 
\def\confName{CVPR}
\def\confYear{2024}

\title{Visual-Augmented Dynamic Semantic Prototype \\for Generative Zero-Shot Learning}

\author{Wenjin Hou$^{1, 4}$, Shiming Chen$^{2 *}$, Shuhuang Chen$^1$, Ziming Hong$^3$, Yan Wang$^4$, Xuetao Feng$^4$, \\Salman Khan$^{2,5}$, Fahad Shahbaz Khan$^{2,6}$, Xinge You$^1$\thanks{Corresponding authors}\\
$^1$Huazhong University of Science and Technology (HUST), China\\
$^{2}$Mohamed bin Zayed University of AI  \quad
$^3$The University of Sydney \\
$\quad$$^4$Alibaba Group \quad $^{5}$Australian National University \quad
$^{6}$Linköping University  \\ 
{\tt\small \{houwj17, gchenshiming\}@gmail.com $\quad$youxg@mail.hust.edu.cn}
}

\begin{document}
\maketitle
\begin{abstract}
    Generative Zero-shot learning (ZSL) learns a generator to synthesize visual samples for unseen classes, which is an effective way to advance ZSL. However, existing generative methods rely on the conditions of Gaussian noise and the predefined semantic prototype, which limit the generator only optimized on specific seen classes rather than characterizing each visual instance, resulting in poor generalizations (\textit{e.g.}, overfitting to seen classes). To address this issue, we propose a novel Visual-Augmented Dynamic Semantic prototype method (termed VADS) to boost the generator to learn accurate semantic-visual mapping by fully exploiting the visual-augmented knowledge into semantic conditions. In detail, VADS consists of two modules: (1) Visual-aware Domain Knowledge Learning module (VDKL) learns the local bias and global prior of the visual features (referred to as domain visual knowledge), which replace pure Gaussian noise to provide richer prior noise information; (2) Vision-Oriented Semantic Updation module (VOSU) updates the semantic prototype according to the visual representations of the samples. Ultimately, we concatenate their output as a dynamic semantic prototype, which serves as the condition of the generator. Extensive experiments demonstrate that our VADS achieves superior CZSL and GZSL performances on three prominent datasets and outperforms other state-of-the-art methods with averaging increases by 6.4\%, 5.9\% and 4.2\% on SUN, CUB and AWA2, respectively.
\end{abstract}    
\section{Introduction}
\label{sec:intro}
\begin{figure*}[t]
  \centering
   \includegraphics[width=0.96\linewidth]{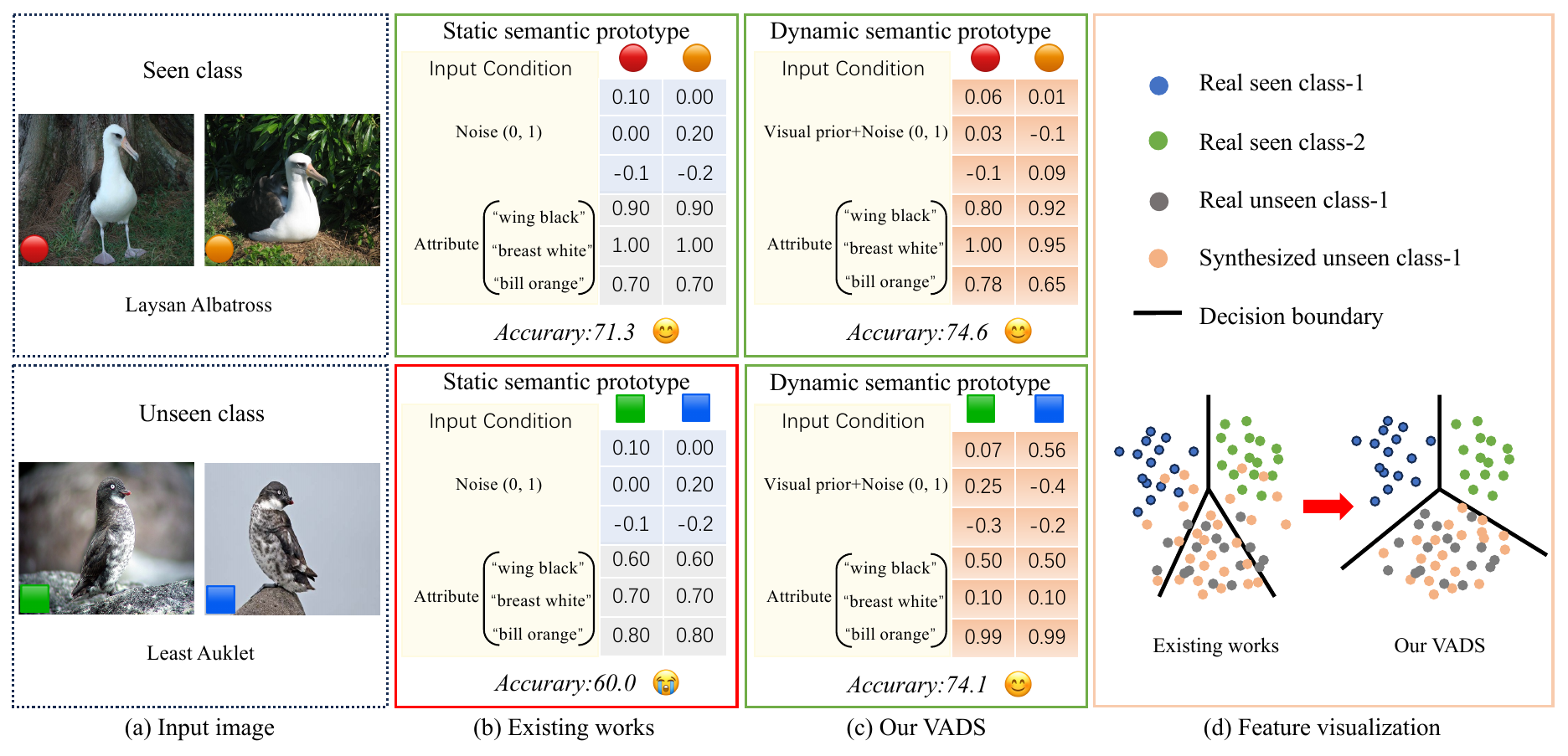}
   \caption{An illustration of the core idea of our method. (a) The semantic prototype (i.e., attribute) of different images of the same category is not fixed, so the predefined semantic prototype is inaccurate in characterizing each instance. (b) Most existing works utilize Gaussian noise and the predefined semantic prototype as conditions to train a semantic$\rightarrow$visual generator on seen classes, which fails to generalize to unseen classes. (c)(d) Our method incorporates rich visual prior with an updated semantic prototype to construct a visual-augmented dynamic semantic prototype of each instance, empowering the generator to synthesize features that faithfully represent the real distribution of unseen classes. Thus, our method achieves better generalization on seen and unseen classes than existing works (\eg, CLSWGAN \cite{xian2018feature}).}
   \label{fig:motivation}
   \vspace{-3mm}
\end{figure*}

Zero-shot learning \cite{palatucci2009zero}, which transfers knowledge from seen classes to unseen classes, has garnered much attention recently. By establishing interactions between visual features and semantic prototypes (also referred to as attribute vectors, side information, or semantic embeddings \cite{xian2019f}), generative ZSL methods exhibit impressive performance, demonstrating the potential of feature synthesis. One of the most successful frameworks is the conditional generative adversarial network (GAN) \cite{goodfellow2014generative}. The main idea of generative ZSL methods is to align semantic prototypes and visual features to synthesize feature of unseen classes. Recent emerging studies have either designed more effective frameworks \cite{xian2018feature,narayan2020latent,chen2021free,han2021contrastive,gupta2023generative} or addressed more specific issues related to visual-semantic alignment \cite{chen2021hsva,chen2023evolving,wang2023improving,cavazza2023no,zhang2024s3a}. These methods have achieved significant improvements.

However, these methods rely on the conditions of Gaussian noise and the predefined semantic prototype (referred to as the static semantic prototype), which limit the generator only optimized on specific seen classes rather than characterizing each visual instance, resulting in poor generalizations (\eg, overfitting to seen classes). \cref{fig:motivation} illustrates these issues: (1) The noise is sampled from a Gaussian distribution $\mathcal N(0,1)$, which lacks the dataset-specific visual prior knowledge (\eg, global visual information ``flying" and ``still" and background information ``sky" and ``grass"). As a result, the domain knowledge shared between seen and unseen classes cannot be utilized for feature synthesis of unseen classes, limiting the knowledge transfer. (2) The predefined semantic prototype fails to characterize each instance well. For example, the attributes ``wing black", ``breast white" and ``bill orange" of the Laysan Albatross are not fixed on different images. Due to these limitations, the visual features synthesized by existing works \cite{xian2018feature,narayan2020latent,chen2022zero,kong2022compactness,hong2022semantic,wang2023improving,cavazza2023no} struggle to represent the distribution of real features, leading to poor generalization to unseen classes, as shown in \cref{fig:motivation}(b). More intuitively, as shown in \cref{fig:motivation}(d), the features of unseen classes synthesized by these methods are confusing, resulting in the decision boundary overfitting to the seen classes. 

Drawing inspiration from image captioning \cite{ramos2023smallcap}, which highlights the generalization of instance-conditional learning, we aim to fully leverage the visual-augmented knowledge into semantic conditions to tackle the aforementioned challenges. On the one hand, we can exploit rich domain visual prior knowledge, serving as a prior noise, to enhance the adaptation and generalization of models \cite{zhou2022conditional, Guo2023ZeroShotGM,yao2023visual}. On the other hand, we can update the predefined semantic prototype to align visual representations based on visual features. As such, incorporating richer and more accurate visual information acts as the semantic condition to train an instance-conditional generative model, which is optimized to characterize each instance (more robust to class shift) rather than to serve only for specific classes. Accordingly, the generative model can synthesize features of unseen classes closer to the real ones, facilitating the classifier in learning an appropriate decision boundary (see the right of \cref{fig:motivation}(d)). 

In this paper, we propose an approach called \underline{\bf V}isual-\underline{\bf A}ugmented \underline{\bf D}ynamic \underline{\bf S}emantic prototype ({\bf VADS}) to improve generative ZSL methods. Specifically, VADS consists of two learnable modules: a Visual-aware Domain Knowledge Learning module (VDKL) and a Vision-Oriented Semantic Updation module (VOSU). The VDKL explores domain visual prior knowledge derived from visual information, which provides richer information for representing instances. The VOSU predicts instance-level semantics through visual$\rightarrow$semantic mapping, guiding the updation of the predefined semantic prototype and promoting accurate semantic prototype learning. Finally, the extracted visual prior and the updated semantic prototype are concatenated as a visual-augmented dynamic semantic prototype, which serves as the condition of the generator during training and feature synthesis, as illustrated in \cref{fig:motivation}(c). Extensive experiments demonstrate the effectiveness of our VADS.

Our contributions can be summarized as follows:

\begin{itemize}
    \item We introduce a Visual-Augmented Dynamic Semantic prototype (VADS) to enhance the generalization of generative ZSL methods, facilitating substantial knowledge transfer. 
   
    \item We devise the VDKL to leverage domain visual prior knowledge from visual features and design the VOSU to dynamically update the predefined semantic prototype. Their outputs together serve as the generator's conditions, providing richer and more accurate visual information.

    \item We conduct extensive experiments on AWA2 \cite{xian2019f}, SUN \cite{patterson2012sun} and CUB \cite{Welinder2010CaltechUCSDB2} datasets. The comprehensive results demonstrate that visual prior knowledge significantly improves the generalization of generative ZSL methods, \textit{i.e.}, average improvements of the harmonic mean over existing generative methods (\eg, f-CLSWGAN \cite{xian2018feature}, TFVAEGAN \cite{narayan2020latent} and FREE \cite{chen2021free}) 6.4\%, 5.9\% and 4.2\% on SUN, CUB and AWA2, respectively. 
    
\end{itemize}

\section{Related Work}
\label{sec:formatting}

\begin{figure*}[t]
  \centering
  \includegraphics[width=16.5cm, height=10.5cm]{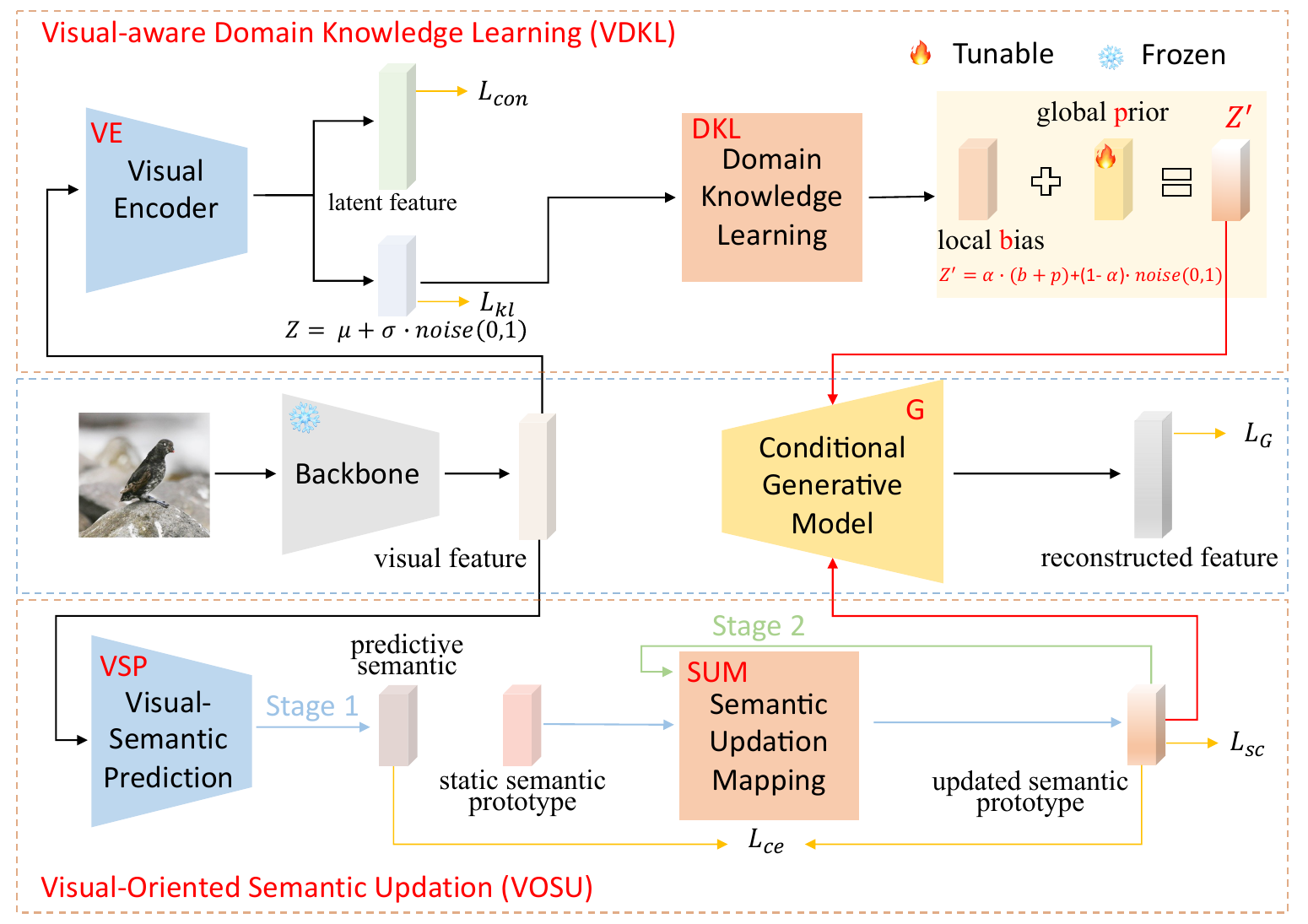}
   \caption{The architecture of our proposed VADS. It consists of two learnable modules: a Visual-Oriented Semantic Updation module (VOSU) and a Visual-aware Domain Knowledge Learning module (VDKL). First, we obtain the prior distribution $\bm Z$ by the Visual Encoder ($\mathit VE$). Following this, the Domain Knowledge Learning network ($\mathit DKL$) transforms $\bm Z$ into a local bias $\bm b$, which is subsequently added to global learnable prior vectors ($\bm p$) to construct the domain visual prior noise (\textit{i.e.}, $\bm {Z^{'}}$). At the bottom, VOSU notably updates the semantic prototype in two stages (depicted by the blue and green arrows). Finally, the visual prior noise and the updated semantic prototype together form a dynamic semantic prototype, used for the reconstruction of features by the generator.} 
   \label{fig:framework}

\end{figure*}
\noindent {\bf Embedding-based Zero-Shot Learning.} Embedding-based ZSL methods are one of the mainstream branches that project visual information into semantic space to align with semantic prototypes. Earlier works \cite{lampert2013attribute,song2018transductive,xie2019attentive} directly mapped global visual features to semantic space, failing to capture local discriminative representation, resulting in suboptimal ZSL performance. Also, embeddings are learned only in seen classes, leading to inevitable seen class bias. In this regard, some studies \cite{huynh2020fine,chen2022transzero,chen2022transzero++} have attempted to use calibration loss to balance the prediction results between seen and unseen classes. Recently, attention mechanisms \cite{vaswani2017attention} have emerged with surprising localization abilities, so semantic-guided methods \cite{xie2019attentive,xu2020attribute,wang2021dual,narayan2021discriminative, chen2022transzero++,chen2022msdn,naeem2022i2dformer,naeem2023i2mvformer} learn to discover attribute-related local regions, providing more accurate inter-class discrimination. Among these methods, APN \cite{xu2020attribute} proposed an attribute prototype network to learn local features, and DPPN \cite{wang2021dual} updated attribute and category prototypes. Inspired by their work, we introduce a dynamic semantic prototype for generative ZSL methods.

\noindent {\bf Generative Zero-Shot Learning.} Generative ZSL methods learn semantic$\rightarrow$visual mapping to synthesize unseen class features, effectively alleviating the lack of unseen class data. Consequently, the quality of synthesized features, which preserves visual-semantic correspondence, plays a crucial role in classification. Thus, TF-VAEGAN \cite{narayan2020latent} forced semantic alignment at all stages, and FREE \cite{chen2021free} fine-tuned visual features to address cross-dataset biases. CE-GZSL \cite{han2021contrastive} and ICCE \cite{kong2022compactness} projected visual features into the latent space for classification. However, these methods constructed projection spaces on seen classes, resulting in inferior generalization ability on unseen classes. Moreover, they uniformly utilize the predefined semantic prototype as a condition, making it difficult to achieve accurate visual-semantic alignment. The method most related to ours is DSP \cite{chen2023evolving}, which updates the prototype by simply adding the evolved and predefined semantic prototype.

\noindent {\bf Large-Scale Vision-Language Models Generalization.} Vision-language models like CLIP \cite{radford2021learning}, pre-trained on large-scale image-text pairs, have demonstrated significant potential for downstream tasks. When performing zero-shot recognition, the class prompts are input into the text encoder to obtain the classification weights, and the cosine similarity between the test image and the weights determines the resulting classification score. It is different from the classical ZSL methods \cite{lampert2013attribute,song2018transductive,xie2019attentive,xu2020attribute,wang2021dual,chen2022transzero++}. Recent research has focused on improving the generalization to unseen classes, with several previous works proposing prompt learning \cite{zhou2022conditional,yao2023visual,abdul2024align}. Motivated by optimizing visual conditional prompts, we introduce visual-aware domain knowledge learning into generative ZSL methods, facilitating knowledge transfer to unseen classes.

\section{Visual-Augmented Dynamic Semantic Prototype Method}

\cref{fig:framework} shows the framework of our VADS. Next, we first present the problem formulation and briefly review the generative ZSL model. Then, we introduce the detailed design of our method.\\
\noindent {\bf Problem Formulation.} Conventional zero-shot learning (CZSL) recognizes unseen classes in the inference stage. Generalized zero-shot learning (GZSL) recognizes both seen and unseen classes. Both settings generalize from seen data $\mathcal{D}^{s}$ to unseen domains $\mathcal{D}^{u}$. $\mathcal {D}^{s}=\{({x}_{i}^s, {y}_{i}^s)|{x}_i^s \in \mathcal {X}^s, {y}_i^s \in \mathcal {Y}^s\}_{i=1}^{N_{s}}$, where $N_s$ is the sample number of seen classes, $x_i^s$ is a feature vector in $\mathcal {X}^s$ and $y_i^s$ is the corresponding lable from $\mathcal {Y}^s$. The $\mathcal{D}^{s}$ is split into a training set $\mathcal{D}_{tr}^{s}$ and a testing set $\mathcal{D}_{te}^s$ following Xian et al. \cite{xian2019f}. Similarly,  $\mathcal {D}^{u}=\{({x}_{i}^u, {y}_{i}^u)|{x}_i^u \in \mathcal {X}^u, {y}_i^u \in \mathcal {Y}^u\}_{i=1}^{N_{u}}$, where $x_i^u$ is a feature vector in $\mathcal {X}^u$ and $y_i^u$ is the lable from $\mathcal {Y}^u$. $\mathcal {Y}^s$ and $\mathcal {Y}^u$ are disjoint. Define attribute semantic prototypes $\mathcal {A}=\mathcal {A}^s \cup \mathcal {A}^u$, corresponding to each category, as a bridge to transfer knowledge from seen classes to unseen classes. In this paper, we dynamically update $\mathcal {A}$ to learn accurate visual-semantic alignment. 

\subsection{Generative ZSL Model}
The goal of the generative ZSL methods is to learn a semantic$\rightarrow$visual generative model ($\mathit G$) on seen classes and then use it to synthesize samples of unseen classes to train a classifier. Existing methods use Gaussian noise and the predefined semantic prototype as input conditions to supervise $\mathit G$ synthetic features (\textit{i.e.}, $\mathcal {A}\times \mathcal {Z}\rightarrow \hat {X}$). In our method, $\mathit G$ represents an off-the-shelf CLSWGAN \cite{xian2018feature}, which contains a generator and a discriminator. We develop the dynamic semantic prototype as a condition, allowing $\mathit G$ to characterize more accurate visual-semantic relationships.

\subsection{Visual-aware Domain Knowledge Learning (VDKL)}
Drawing inspiration from previous prompt learning \cite{zhou2022conditional,yao2023visual}, we exploit the rich information in visual features to assist in synthesizing features. VDKL is a data-efficient module allowing the visual features to be used to improve generalization. As shown in \cref{fig:framework}, we design a Visual Encoder ($\mathit {VE}$) and a Domain Knowledge Learning network ($\mathit {DKL}$). First, the $\mathit {VE}$ encodes visual features into a latent feature $l$ and a latent code $z$. The latent feature enables inter-class alignment of visual features, and latent code is subsequently confined to a prior distribution $\bm Z$. The optimization of the $\mathit {VE}$ is achieved via contrastive loss \cite{chen2020simple} and evidence-lower bound given by the equation as follows:

\begin{equation}
  \label{eq:L_con}
  \mathcal{L}_{con} = \mathbb{E} [ \mathrm {log}\frac{\mathrm{exp}(l_{i}^{T}l^{+}/\tau) }{\mathrm{exp}(l_{i}^{T}l^{+}/\tau)+\sum_{k=1}^{K} \mathrm{exp}(l_{i}^{T}l_{k}^{-}/\tau) } ],
\end{equation}
\begin{equation}
  \label{eq:L_kl}
  \mathcal{L}_{kl} =  KL({VE(x)}||p(z)),
\end{equation}
where $l^{+}$ and $l_{k}^{-}$ represent positive and negative latent features, $\tau$ is a temperature parameter set as 0.15, $K$ is the class numbers, ${KL}$ denotes the Kullback-Leibler divergence and $p(z)$ is a prior distribution that is assumed to be $\mathcal {N}(0, 1)$.

To further utilize visual prior knowledge during the training and synthesis stages, we propose a Domain Knowledge Learning network ($\mathit {DKL}$) to obtain a local bias $\bm b$ of visual features (\textit{i.e.}, $\bm {b} = DKL(\bm{Z})$). Additionally, we employ a learnable prior vector $\bm p$ to capture global visual information ($\bm p$ is randomly initialized). Subsequently, we obtain domain-specific visual prior noise as follows:
\begin{align}
    \label{eq:p}
    \bm {Z^{'}} = \alpha\cdot(\bm{b}+\bm{p})+(1-\alpha)\cdot \bm{noise(0,1)},
  \end{align}

  where the $\bm {noise(0,1)}$ represents Gaussian noise aimed at enhancing diversity in synthesis, $\alpha$ is the combination coefficient set as 0.9. Through this operation, we argue that $\bm{Z^{'}}$ includes rich domain visual knowledge and feeds it into the generator to provide instance conditions, promoting the generator learning and utilizing it for feature synthesis of unseen classes. Note that unseen class samples are unavailable in the feature synthesis stage, so we randomly sample Gaussian noise input to $\mathit {DKL}$, transferring the domain knowledge acquired from seen classes to unseen classes.
\subsection{Visual-Oriented Semantic Updation (VOSU)}
We observe that the predefined semantic prototype struggles to represent each visual sample accurately, so we propose a Visual-Oriented Semantic Updation module (VOSU), optimizing the semantic prototype dynamically. Our semantic prototype updation involves a two-stage process. In the first stage, we feed visual features $x_s$ into the Visual-Semantic Prediction network ($\mathit {VSP}$) to generate a predictive semantic $\hat{a}$ that explicitly captures specific visual patterns of the target image. Then, the predefined semantic prototype is input into a Semantic Updation Mapping network ($\mathit {SUM}$) to learn an updated semantic $\dot{a}$. This mapping can be expressed as:
\begin{align}
    \label{eq:ua}
  \dot{a} = {SUM}(a).
  \end{align}

To maintain the attribute information of the prototype and integrate the visual information, we jointly optimize them by the cross-entropy loss $\mathcal {L}_{ce}$. $\mathcal {L}_{ce}$ is defined as:
\begin{align}
  \label{eq:lce}
 \mathcal {L}_{ce} = -\frac{1}{N} \sum_{i=1}^{N} \mathrm {log}\frac{\mathrm{exp}({VSP}(x_{i})^T \dot{a}^{y})}{ \sum_{\bar{c}\in C^{s\cup u}}\mathrm {exp}({VSP}(x_{i})^T \dot{a}^{\bar{c}}) },
\end{align}
where $N$ is the batch numbers. Accordingly, the updated semantic prototype incorporates rich visual information. In the second stage, we employ $\mathit{SUM}$ to update $\dot{a}$ during the conditional generative model training and use it as a condition to learn together with the generator $\mathit{G}$ and the discriminator $\mathit{D}$. This implementation facilitates dynamic updation and accurate visual-semantic matching. To this end, we propose the semantic consistency loss $\mathcal {L}_{sc}$ as follows:
\begin{gather}
  \label{eq:L_ce}
  \mathcal{L}_{sc}= \mathbb{E}\left[\|\mathrm {\mathit {SUM}}(\dot{a})-\dot{a}\|_{1}\right].
\end{gather}

In summary, the first stage leverages visual features to assist semantic updation, and the second stage dynamically updates the prototype of each sample. Then, we concatenate the updated semantic prototype with the visual prior noise $Z^{'}$, called the dynamic semantic prototype, which serves as the condition for the generator, as depicted in \cref{fig:framework}.

\subsection{Overall Objective and Inference}
\noindent {\bf VADS Objective Loss Function.} Overall, the objective loss function of VADS is:
\begin{gather}
\label{eq:L_total}
\mathcal{L}_{total}= \mathcal{L}_{G}+ \lambda_{con}\mathcal{L}_{con}+\lambda_{kl}\mathcal{L}_{kl}+\lambda_{sc}\mathcal{L}_{sc}, 
\end{gather}
where $\mathcal{L}_{G}$ is the loss of conditional generative model G, $\lambda_{con}$, $\lambda_{kl}$ and $\lambda_{sc}$ are the hyper-parameter to balance each loss term. To fully validate our method, by using this loss, we train on various mainstream generative models (\eg, CLSWAGN \cite{xian2018feature}, TFVAEGAN \cite{narayan2020latent}, and FREE \cite{chen2021free}). Next, we illustrate feature synthesis and classifier training.\\

\noindent {\bf Visual-Augmented Feature Synthesis for Unseen Classes.} To fully utilize the visual knowledge and accurate semantic prototypes, we sample Gaussian noise input $\mathit{DKL}$ to obtain the prior noise $\bm{Z^{'}}$ (\textit{i.e.,} \cref{eq:p})and use $\mathit {SUM}$ to update the semantic prototypes of unseen classes (\textit{i.e.}, $\dot{a}_{u}=SUM(\hat{a}_{u})$). They serve as conditions to synthesize visual samples closer to the real features for training the classifier. The form can be written as:
\begin{align}
  \label{eq:syn}
\hat{x}_{u} = \mathrm {G}(\bm{Z^{'}},\dot{a}_u),
\end{align}
where $\dot{a}_u$ is updated semantic prototype of the unseen classes and $\hat{x}_{u}$ is the synthesized features of unseen classes.\\

\noindent {\bf ZSL Classifier Training and Inference.} After synthesizing features, we input the seen class training features and synthesized unseen class features into $\mathit {VE}$ to extract latent features and concatenate them to enhance the original features, alleviating cross-dataset bias \cite{chen2021free}. Then, we train a CZSL classifier using enhanced-synthetic features (i.e., $f_{CZSL}: \mathcal{X} \rightarrow  \mathcal {Y}^u$) and train a GZSL classifier using enhanced seen class training features and enhanced-synthetic features (i.e., $f_{GZSL}: \mathcal{X} \rightarrow \mathcal{Y}^{s} \cup \mathcal{Y}^{u}$). Finally, we perform inference using the test sets $\mathcal {D}_{te}^{s}$ and $\mathcal {D}^{u}$.

\section{Experiments}
\begin{table*}[ht]
  \centering
  \caption{Compared our VADS with the state-of-the-art on AWA2, SUN and CUB benchmark datasets in the CZSL and GZSL settings. The best and second-best results are marked in \textbf{\color{red}Red} and \textbf{\color{blue}Blue}, respectively. Symbol ``--" denotes no results are reported.}
  \resizebox{\linewidth}{!}{
  \begin{tabular}{c|l|c|c|cccc|cccc|cccc}
      \hline
      \multirow{3}{*}{\textbf{Type}}&\multirow{3}{*}{\textbf{Methods}}  &\multirow{3}{*}{\textbf{Venue}}&\multirow{3}{*}{\textbf{Backbone}}&\multicolumn{4}{c|}{\textbf{AWA2}}&\multicolumn{4}{c|}{\textbf{SUN}}&\multicolumn{4}{c}{\textbf{CUB}}\\
      \cline{5-16}
      &&&&\multicolumn{1}{c|}{CZSL} & \multicolumn{3}{c|}{GZSL}&\multicolumn{1}{c|}{CZSL} & \multicolumn{3}{c|}{GZSL}&\multicolumn{1}{c|}{CZSL} & \multicolumn{3}{c}{GZSL}\\
      \cline{5-16} 
      &&&&\multicolumn{1}{c|}{\rm{Acc}} & \rm{U} & \rm{S} &\rm{H} & \multicolumn{1}{c|}{\rm{Acc}} & \rm{U} &\rm{S} & \rm{H} & \multicolumn{1}{c|}{\rm{Acc}} &\rm{U}  & \rm{S}  & \rm{H}\\
      \hline
      \multirow{9}{*}{{\rotatebox{90}{Embedding}}}
      & PREN \cite{ye2019progressive} &CVPR'19&ResNet-101&\multicolumn{1}{c|}{\color{blue}\bf {74.1}}&32.4 &88.6 &47.4&\multicolumn{1}{c|}{62.9}&35.4& 27.2 &30.8& \multicolumn{1}{c|}{66.4} &  35.2 &55.8& 43.1\\
      & DAZLE \cite{huynh2020fine}&CVPR'20 &ResNet-101&\multicolumn{1}{c|}{67.9}&60.3& 75.7&67.1&\multicolumn{1}{c|}{59.4} &52.3 &24.3 &33.2&\multicolumn{1}{c|}{66.0} &56.7&59.6&58.1\\
      & DVBE \cite{min2020domain}&CVPR'20&ResNet-101 &\multicolumn{1}{c|}{--}&63.6& 70.8& 67.0&\multicolumn{1}{c|}{--}& 45.0& 37.2& 40.7&\multicolumn{1}{c|}{--}& 53.2& 60.2& 56.5\\
      &CN \cite{skorokhodov2020class}&ICLR'21&ResNet-101 &\multicolumn{1}{c|}{--}&60.2&77.1&{67.6}&\multicolumn{1}{c|}{--}&44.7&41.6&43.1&\multicolumn{1}{c|}{--}&49.9&50.7&50.3\\
      &GEM-ZSL \cite{liu2021goal} & CVPR'21&ResNet-101&\multicolumn{1}{c|}{67.3}&64.8&77.5&70.6&\multicolumn{1}{c|}{62.8}&38.1&35.7&36.9&\multicolumn{1}{c|}{77.8}&64.8&77.1&70.4\\
      &ViT-ZSL \cite{alamri2021multi}&IMVIP'21&ViT-Large&\multicolumn{1}{c|}{--}&51.9&{\color{blue}\bf 90.0}&65.8&\multicolumn{1}{c|}{--}&44.5&{\color{red}\bf 55.3}&49.3&\multicolumn{1}{c|}{--}&67.3&75.2&71.0\\
      &IEAM-ZSL \cite{alamri2021implicit} &DGAM'21&ViT-Large&\multicolumn{1}{c|}{--}&53.7&89.9&67.2&\multicolumn{1}{c|}{--}&48.2&{\color{blue}\bf 54.7}&51.3&\multicolumn{1}{c|}{--}&68.6&73.8&71.1\\
      & DUET \cite{chen2023duet}&AAAI'23&ViT-Base&\multicolumn{1}{c|}{69.9}&63.7&84.7&72.7&\multicolumn{1}{c|}{64.4}&45.7&45.8&45.8&\multicolumn{1}{c|}{72.3}&62.9&72.8&67.5\\
      &PSVMA \cite{liu2023progressive}&CVPR'23&ViT-Base&\multicolumn{1}{c|}{--}&\color{blue}\bf{73.6}&77.3&{\color{blue}\bf 75.4}&\multicolumn{1}{c|}{--}&{\color{blue}\bf 61.7}&45.3&{\color{blue}\bf 52.3}&\multicolumn{1}{c|}{--}&{\color{blue}\bf 70.1}&{\color{blue}\bf 77.8}&{\color{blue}\bf 73.8}\\ 
      \hline
      \multirow{12}{*}{{\rotatebox{90}{Generative}}}

      & f-VAEGAN-D2 \cite{xian2019f} &CVPR'19  &ResNet-101   &\multicolumn{1}{c|}{71.1}&57.6& 70.6&63.5&\multicolumn{1}{c|}{64.7} &45.1 &38.0 &41.3&\multicolumn{1}{c|}{61.0} &48.4&60.1&53.6\\
      & TF-VAEGAN \cite{narayan2020latent}&ECCV'20  &ResNet-101  &\multicolumn{1}{c|}{72.2}&59.8& 75.1&66.6&\multicolumn{1}{c|}{\color{blue}\bf 66.0} &45.6 &40.7&43.0&\multicolumn{1}{c|}{64.9} &52.8&64.7&58.1\\
      & FREE \cite{chen2021free}  &ICCV'21&ResNet-101  &\multicolumn{1}{c|}{--}&60.4& 75.4&67.1&\multicolumn{1}{c|}{--} &47.4 &37.2&41.7&\multicolumn{1}{c|}{--} &55.7&59.9&57.7\\
      & HSVA \cite{chen2021hsva} &NeurIPS'21 & ResNet-101  &\multicolumn{1}{c|}{--}&56.7& 79.8&66.3&\multicolumn{1}{c|}{--} &48.6 &39.0&43.3&\multicolumn{1}{c|}{--} &52.7&58.3&55.3\\
      & CE-GZSL \cite{han2021contrastive}&CVPR'21 & ResNet-101 &\multicolumn{1}{c|}{70.4} &63.1&78.6&70.0&\multicolumn{1}{c|}{63.3}&48.8&38.6&43.1&\multicolumn{1}{c|}{77.5}&63.9&66.8&65.3\\
      & FREE+ESZSL \cite{cetin2022closed} &ICLR'22 &ResNet-101 &\multicolumn{1}{c|}{--}&51.3& 78.0&61.8   &\multicolumn{1}{c|}{--} &48.2 &36.5&41.5&\multicolumn{1}{c|}{--} &51.6&60.4&55.7\\
      &ICCE \cite{kong2022compactness}&CVPR'22&ResNet-101&\multicolumn{1}{c|}{72.7}&65.3&82.3&72.8&\multicolumn{1}{c|}{--}&--&--&--&\multicolumn{1}{c|}{\color{blue}\bf 78.4}&67.3&65.5&66.4\\
      &TDCSS \cite{feng2022non}&CVPR'22&ResNet-101&\multicolumn{1}{c|}{--}&59.2&74.9&66.1&\multicolumn{1}{c|}{--}&--&--&--&\multicolumn{1}{c|}{--}&44.2&62.8&51.9\\
      &CDL + OSCO \cite{cavazza2023no}&TPAMI'23&ResNet-101&\multicolumn{1}{c|}{--}&48.0&71.0&57.1&\multicolumn{1}{c|}{--}&32.0&65.0&42.9&\multicolumn{1}{c|}{--}&29.0&69.0&40.6\\
      &CLSWGAN+DSP \cite{chen2023evolving}&ICML'23 &ResNet-101&\multicolumn{1}{c|}{--}&60.0&86.0&70.7& \multicolumn{1}{c|}{--}&48.3&43.0&45.5&\multicolumn{1}{c|}{--}&51.4&63.8&56.9\\ 
      &TFVAEGAN+SHIP \cite{wang2023improving}&ICCV'23&ViT-Base&\multicolumn{1}{c|}{--}&61.2&{\color{red}\bf 95.9}&74.7&\multicolumn{1}{c|}{--}&--&--&--&\multicolumn{1}{c|}{--}&22.5&{\color{red}\bf 82.2}&35.3\\
      \cline{2-16} 
      &\bf{VADS (Ours)} & \multicolumn{1}{c|}{--}&ViT-Base&\multicolumn{1}{c|}{\color{red}\bf {82.5}}&{\color{red}\bf {75.4}}&83.6&{\color{red}\bf79.3}&\multicolumn{1}{c|}{{\color{red}\bf 76.3}}&{\color{red}\bf 64.6}&49.0&{\color{red}\bf 55.7}&\multicolumn{1}{c|}{\color{red}\bf 86.8}&{\color{red}\bf 74.1}&74.6&{\color{red}\bf 74.3}\\
      
      \hline
  \end{tabular}}
  
  \label{tab:sota}
\end{table*}
\subsection{Experimental Setup}
\noindent {\bf Benchmark Datasets.} We conduct extensive experiments on three prominent ZSL benchmark datasets: Animals with Attributes 2 (AWA2 \cite{xian2019f}), SUN Attribute (SUN \cite{patterson2012sun}) and Caltech-USCD Birds-200-2011 (CUB \cite{Welinder2010CaltechUCSDB2}). We follow the Proposed Split (PS) setting \cite{xian2019f} to split each dataset into seen and unseen classes, as detailed in \cref{tab:dataset}.
\begin{table}[ht]
  \centering
  \caption{A detailed illustration of the ZSL benchmark datasets. $s$ and $u$ are the number of seen and unseen classes. $N_{A}$ refers to semantic prototype dimensions. We follow CE-GZSL \cite{han2021contrastive} using 1024-dimensional semantic prototypes generated from textual descriptions \cite{reed2016learning} on CUB.
  }
      \begin{tabular}{l|c|c|c}
        \hline
        Datasets &  \# images &  \# classes ($s$ $|$ $u$) & \# $N_{A}$ \\
        
        \hline
        {\bf AWA2} \cite{xian2019f}& 37322&50 (40 $|$ 10)&85\\
        {\bf SUN} \cite{patterson2012sun} & 14340 & 717 (645 $|$ 72) & 102 \\
        {\bf CUB} \cite{Welinder2010CaltechUCSDB2} & 11788 & 200 (150 $|$ 50) & 1024\\
        \hline
    \end{tabular}
      \label{tab:dataset}
      \vspace{-4mm}
\end{table}

\noindent {\bf Evaluation Protocols.} During inference (\textit{i.e.}, performing CZSL and GZSL classification), we follow the evaluation protocols in \cite{xian2019f}. In the CZSL setting, we calculate the average per-class Top-1 accuracy of unseen classes, denoted as $\bm {Acc}$. For the GZSL scenario, we measure the Top-1 accuracy of the seen and unseen classes, defined as $\bm S$ and $\bm U$, respectively. We also compute the harmonic mean, defined as $\bm{H} = (2 \times \bm{S}\times \bm{U})/(\bm{S}+\bm{U})$. \\
\noindent {\bf Implementation Details.} We follow PSVMA \cite{liu2023progressive} using the ViT-Base Backbone \cite{dosovitskiy2020image} without fine-tuning as the feature extractor, obtaining 768-dimensional visual features for all samples. The global prior $\bm p$ has the same dimension as the semantic prototype. We set the mini-batch to 64, 128 and 128 for AWA2, SUN and CUB, respectively. We use the Adam optimizer \cite{kingma2014adam} with $\beta_1$ = 0.5, $\beta_2$ = 0.999,  and set the initial learning rate to 0.0001. We synthesize 5600, 100, and 400 samples for each class on AWA2, SUN and CUB. Our experiments are based on the PyTorch and implemented on a NVIDIA GeForce RTX 3090 GPU.

\subsection{Comparison with State-of-the-Art Methods} 
We report the performance of our proposed VADS using CLSWGAN as a generative model compared to state-of-the-art methods. \cref{tab:sota} shows the results, including the embedding-based and generative ZSL methods. In the CZSL scenario, our method notably outperforms the sub-optimal results by 8.4\%, 10.3\%, and 8.4\%, achieving the best results on AWA2, SUN, and CUB. The results confirm that our method incorporating dynamic semantic prototypes is more generalizable to unseen classes than static semantic prototypes. In the GZSL scenario, our method obtains the best harmonic mean $\bm H$ on all datasets (\textit{i.e.}, AWA2 ($\bm H=$ 79.3), SUN ($\bm H=$ 55.7) and CUB ($\bm H=$ 74.3)). Our method significantly outperforms CLSWGAN+DSP \cite{chen2023evolving}, which proposed an evolved semantic prototype, indicating that our design is reasonable. Meanwhile, our method is also competitive compared to the recent PSVMA \cite{liu2023progressive}, DUET \cite{chen2023duet}, and TFVAEGAN+SHIP \cite{wang2023improving} methods using the ViT Backbone. Furthermore, our method achieves optimal results in unseen class accuracy $\bm U$, demonstrating that the features synthesized by the generator are closer to the real features of unseen classes, effectively alleviating the over-fitting problem. Noted that TFVAEGAN+SHIP \cite{wang2023improving} using CLIP Encoder and ViT-ZSL \cite{alamri2021implicit} using ViT-Large achieve the best and second-best accuracy for seen classes, but they fail to generalize well to unseen classes. These results consistently demonstrate that our method synthesizes reliable features of unseen classes to facilitate classifier learning, resulting in superior ZSL performance.

\begin{table}[htbp]
  \small
  \centering
  \caption{Ablation study of VADS on modules, feature enhancement and loss terms on CUB and AWA2. We use CLSWGAN \cite{xian2018zero} as a generative model. The best result is marked in \textbf{boldface}.}
  \resizebox{0.45\textwidth}{!}{
      \begin{tabular}{l|c|c|c|c}
          \hline
          \multirow{2}*{Configurations} &\multicolumn{2}{c|}{\textbf{CUB}} &\multicolumn{2}{c}{\textbf{AWA2}}\\
          \cline{2-5}
          & \rm{Acc} & \rm{H}&\rm{Acc} & \rm{H} \\
          \hline
          (1) VADS w/o VDKL \& VOSU &80.1&65.2	&71.3&69.3\\ 
          (2) VADS w/o VDKL &84.9&72.8	&78.1&	78.5\\
          (3) VADS w/o VOSU &83.8	&70.4	&75.4&	77.0\\
          (4) VADS w/o enhancement &85.1&73.4&81.8&79.1\\ 
          (5) VADS w/o $\mathcal{L}_{con}$ (\textit{i.e.}, \cref{eq:L_con}) &85.3&73.1&79.4&78.6\\
          (6) VADS w/o $\mathcal{L}_{sc}$ (\textit{i.e.}, \cref{eq:L_ce})&86.0&73.5&79.5&78.7\\
          \hline
          (7) \bf{VADS (full)}&\bf{86.8}&\bf{74.3}&\bf{82.5}&\bf{79.3}\\
          
          \hline
      \end{tabular}
      }

      \label{tab:ablation}
      \vspace{-6mm}
\end{table}
\subsection{Ablation Study}

\begin{table*}[ht]
  \centering
  \caption{Evaluation of VADS with multiple generative ZSL models on three prominent datasets using ViT-Base Backbone. Each row pair shows the effect of adding VADS to a particular generative ZSL model. We use the same hyperparameter optimization policy in all cases to make results comparable.}
  \resizebox{\linewidth}{!}{\small
  \begin{tabular}{l|cccc|cccc|cccc}
      \hline
      \multirow{3}{*}{\textbf{Generative ZSL Methods}}  &\multicolumn{4}{c|}{\textbf{AWA2}}&\multicolumn{4}{c|}{\textbf{SUN}}&\multicolumn{4}{c}{\textbf{CUB}}\\
      \cline{2-13}
      &\multicolumn{1}{c|}{CZSL} & \multicolumn{3}{c|}{GZSL}&\multicolumn{1}{c|}{CZSL} & \multicolumn{3}{c|}{GZSL}&\multicolumn{1}{c|}{CZSL} & \multicolumn{3}{c}{GZSL}\\
      \cline{2-13}
      &\multicolumn{1}{c|}{\rm{Acc}} & \rm{U} & \rm{S} &\rm{H} & \multicolumn{1}{c|}{\rm{Acc}} & \rm{U} &\rm{S} & \rm{H} & \multicolumn{1}{c|}{\rm{Acc}} &\rm{U}  & \rm{S}  & \rm{H} \\
      \hline
      CLSWGAN \cite{xian2018feature}&\multicolumn{1}{c|}{71.3}& 66.2& 72.6&69.3 & \multicolumn{1}{c|}{66.0} &46.9&42.6&44.6& \multicolumn{1}{c|}{80.1} &60.0&71.3&65.2\\
      CLSWGAN + \textbf{VADS} &\multicolumn{1}{c|}{82.5$^{\color{blue}\text{+11.5}}$}&75.4&83.6&79.3$^{\color{blue}\text{+10.0}}$&\multicolumn{1}{c|}{76.3$^{\color{blue}\text{+10.3}}$}&64.6&49.0& 55.7$^{\color{blue}\text{+11.1}}$&\multicolumn{1}{c|}{ 86.8$^{\color{blue}\text{+6.7}}$}& 74.1&74.6& 74.3$^{\color{blue}\text{+9.1}}$\\
      \hline
      TFVAEGAN \cite{narayan2020latent}     &\multicolumn{1}{c|}{78.2} &66.7&87.1&75.6&\multicolumn{1}{c|}{73.1} &60.6 &48.6 &54.0&\multicolumn{1}{c|}{81.6}&64.8& 74.6&69.3\\
      TFVAEGAN + \textbf{VADS} &\multicolumn{1}{c|}{80.2${^{\color{blue} \text{+2.0}}}$}&75.7&	83.3	&79.3${^{\color{blue}\text{+3.7}}}$&\multicolumn{1}{c|}{76.3${^{\color{blue}\text{+3.2}}}$}&61.9	&51.0&	55.9${^{\color{blue}\text{+1.9}}}$&\multicolumn{1}{c|}{83.6${^{\color{blue} \text{+2.0}}}$}&70.1&70.9&70.5${^{\color{blue}\text{+1.2}}}$\\
      \hline
      FREE \cite{chen2021free}     &\multicolumn{1}{c|}{70.6} &62.9&	85.9&	72.6&\multicolumn{1}{c|}{71.7} &45.4&	50.4	&47.8&\multicolumn{1}{c|}{84.3}&68.7&	73.5	&70.9\\
      FREE + \textbf{VADS} &\multicolumn{1}{c|}{79.4${^{\color{blue}\text {+8.8}}}$}&70.1	&84.6	&76.6${^{\color{blue}\text{+4.0}}}$&\multicolumn{1}{c|}{75.0${^{\color{blue}\text{+3.3}}}$}&57.6	&50.7	&53.9${^{\color{blue}\text{+6.1}}}$&\multicolumn{1}{c|}{85.5${^{\color{blue}\text{+1.2}}}$}&70.9	&75.4	&73.1${^{\color{blue}\text{+3.2}}}$\\
      
      \hline
  \end{tabular}}

  \label{tab:com}
  \vspace{-3mm}
\end{table*}

\noindent {\bf Component Analysis.}
In this section, we perform a series of experiments to analyze the effectiveness of significant components. \cref{tab:ablation} summarizes the results of ablation studies on CUB and AWA2. We first use ViT-Base to extract visual features to train CLSWAGN as the baseline. Compared to the baseline ($\textit{i.e.}$, configuration (1)), our VADS (configuration (7)) performance improves significantly (\textit{i.e.}, the ${\bm{Acc/H}}$ increases by 6.7\%/9.1\% on CUB and 11.5\%/10\% on AWA2). Configurations (2) and (3) correspond to the effectiveness of the  VDKL and VOSU modules. When there is no VDKL, that is, no visual prior is introduced to generate samples, the performance drops by 1.9\%/1.5\% and 4.4\%/0.8\% for CUB and AWA2, indicating that visual prior is beneficial for transferring knowledge to unseen classes. When without VOSU, the performance drops most severely, verifying visual-semantic alignment is crucial for learning an accurate generator $\mathit G$. Next, without $\mathit {VE}$ enhancing classification features, the results of configuration (4) drop slightly in both CZSL and GZSL settings, which shows that feature enhancement mitigates cross-dataset bias. Lastly, we conduct experiments to study the impact of the loss terms on performance. Configuration (5) without contrastive loss $\mathcal{L}_{con}$, $\bm{Acc/H}$ decreases by 1.5\%/1.2\% and 3.1\%/0.7\% on CUB and AWA2, respectively. The main reason is that contrastive loss achieves feature alignment, without which the network may learn category-agnostic redundant information. For the lack of semantic consistency loss $\mathcal {L}_{sc}$ (\textit{i.e.}, configuration (6)), there is no guarantee that the semantic prototype maintains the original definition when training on seen classes, thus hurting the performance of seen classes. Note that the cross-entropy loss $\mathcal {L}_{ce}$ and KL loss $\mathcal {L}_{kl}$ are necessary for single network training, so we did not perform ablation. 
\\
\begin{figure}[t]
  \centering
  
   \includegraphics[width=1.0\linewidth]{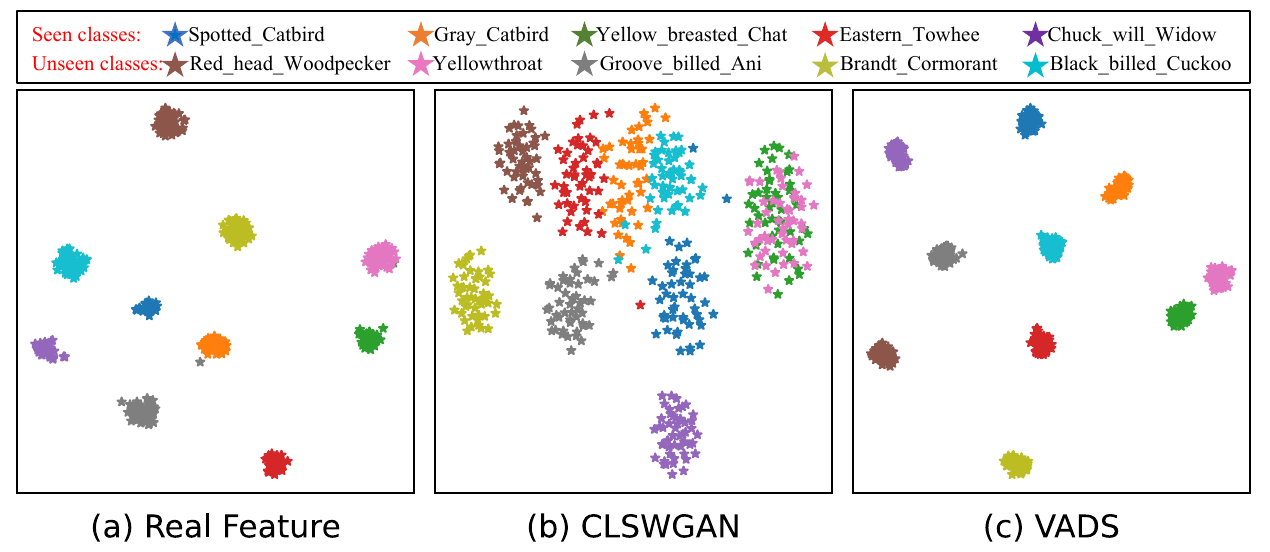}
   \caption{t-SNE visualizations on CUB. The 10 different colors refer to the 5 seen classes and 5 unseen classes that are randomly selected. Please zoom in for a better view.}
   \label{fig:tsne_cub}
   \vspace{-4mm}
\end{figure}

\noindent {\bf Qualitative Evaluation.} To further demonstrate that our method synthesizes reliable unseen class features for classifier training, we project various visual features into two principal components via t-SNE \cite{van2008visualizing}. \cref{fig:tsne_cub} shows the t-SNE visualization on CUB. Each color of '$\star$' represents a class, and we randomly select 5 seen classes and 5 unseen classes. From left to right, they represent the real visual features extracted by ViT-Base \cite{dosovitskiy2020image}, the features synthesized by CLSWGAN, and the features synthesized by our VADS. We observe that the real visual features are inter-class dispersion and intra-class aggregation. In \cref{fig:tsne_cub}(b), Gaussian noise and the predefined semantic prototype serve as conditions to synthesize samples by CLSWGAN. There are two apparent phenomena: first, the feature distributions are scattered, which cannot truly reflect each class; second, the synthesized seen and unseen class features are confusing (\eg, ``{\color{green}{$\star$}}" and ``{\color{pink}{$\star$}}", which denote seen class ``Yellow breasted Chat" and unseen class ``Yellowthroat", respectively). Therefore, the decision boundary of the CZSL/GZSL classifier trained with these features is unclear, consistent with the motivation of \cref{fig:framework}(d). In contrast, the features generated with our VADS are closer to the real features and are inter-class separated, as shown in \cref{fig:tsne_cub}(c). On the one hand, we analyze that visual-augmented dynamic semantic prototype motivates the generator to learn accurate semantic$\rightarrow$visual mapping. On the other hand, synthesized unseen class features are more reliable, leading to learning appropriate classification boundaries of unseen classes.

\subsection{Generative ZSL Models with VADS} 
To further evaluate VADS as a generic technology to improve generative ZSL, we integrate it into three prevalent generative ZSL frameworks: CLSWGAN \cite{xian2018zero}, TF-VAEGAN \cite{narayan2020latent}, and FREE \cite{chen2021free}. We use the official repository to reproduce the results and then insert our module to verify our method's effectiveness. Note that TF-VAEGAN and FREE contain a visual encoder, so we maintain their design. When inserting our modules, we keep the hyper-parameters unchanged to make the results comparable. The results on the three datasets are presented in \cref{tab:com}.  In terms of $\bm {Acc}$ and $\bm {H}$, we observe varying degrees of performance improvements (\eg, a maximum of 11.5 points and a minimum of 1.2 points). Average growth is 7.4\%/5.9\%, 5.6\%/6.4\% and 3.3\%/4.2\% for ${\bm{Acc/H}}$ on AWA2, SUN and CUB. Overall, the consistent improvement over competitive benchmarks validates the effectiveness of our proposed method.
\begin{figure*}[ht]
  \begin{center}
    \begin{tabular}{cccc}
      \hspace{-1mm}\includegraphics[width=0.22\linewidth]{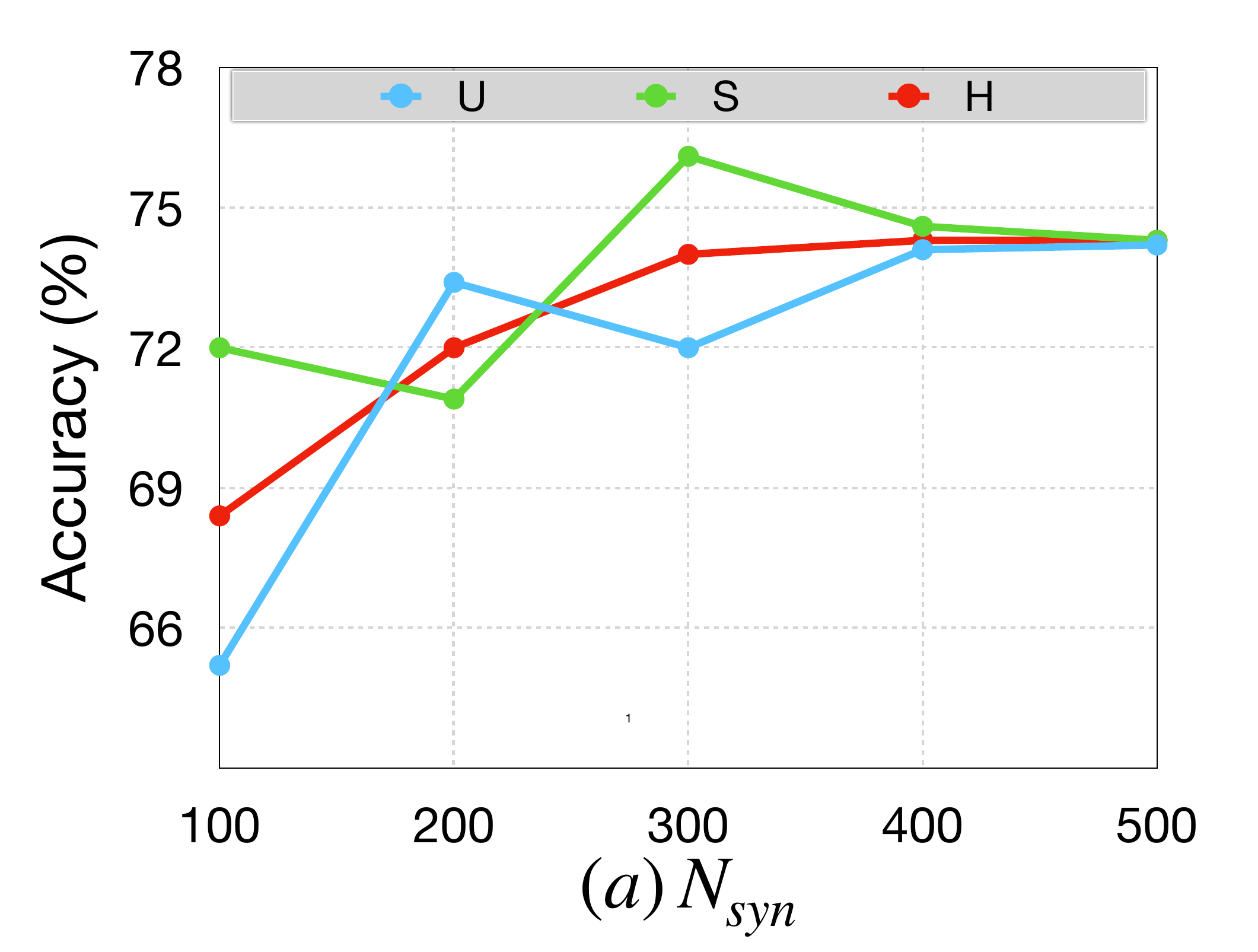} &
      \hspace{-5mm}\includegraphics[width=0.22\linewidth]{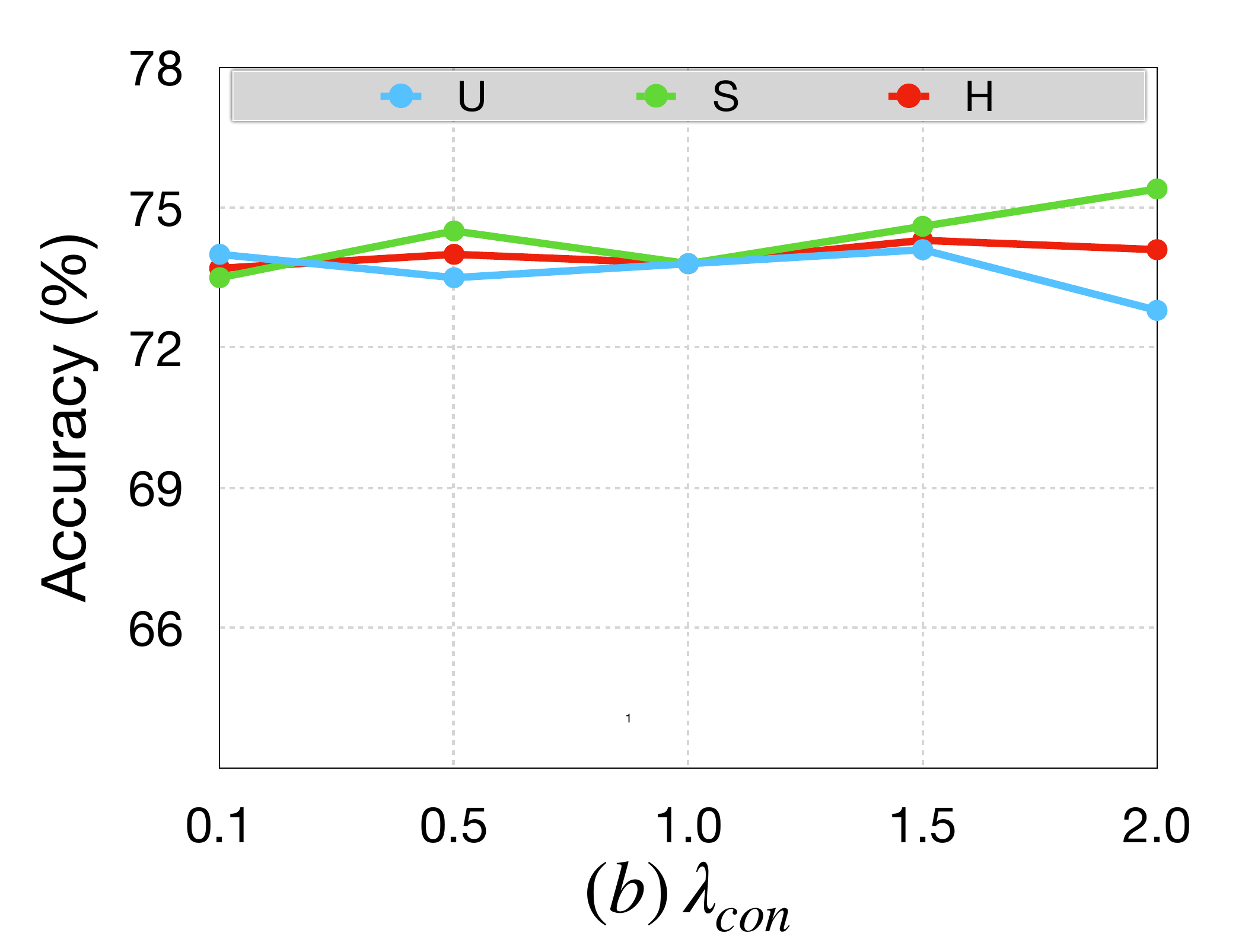}&
      \hspace{-5mm}\includegraphics[width=0.22\linewidth]{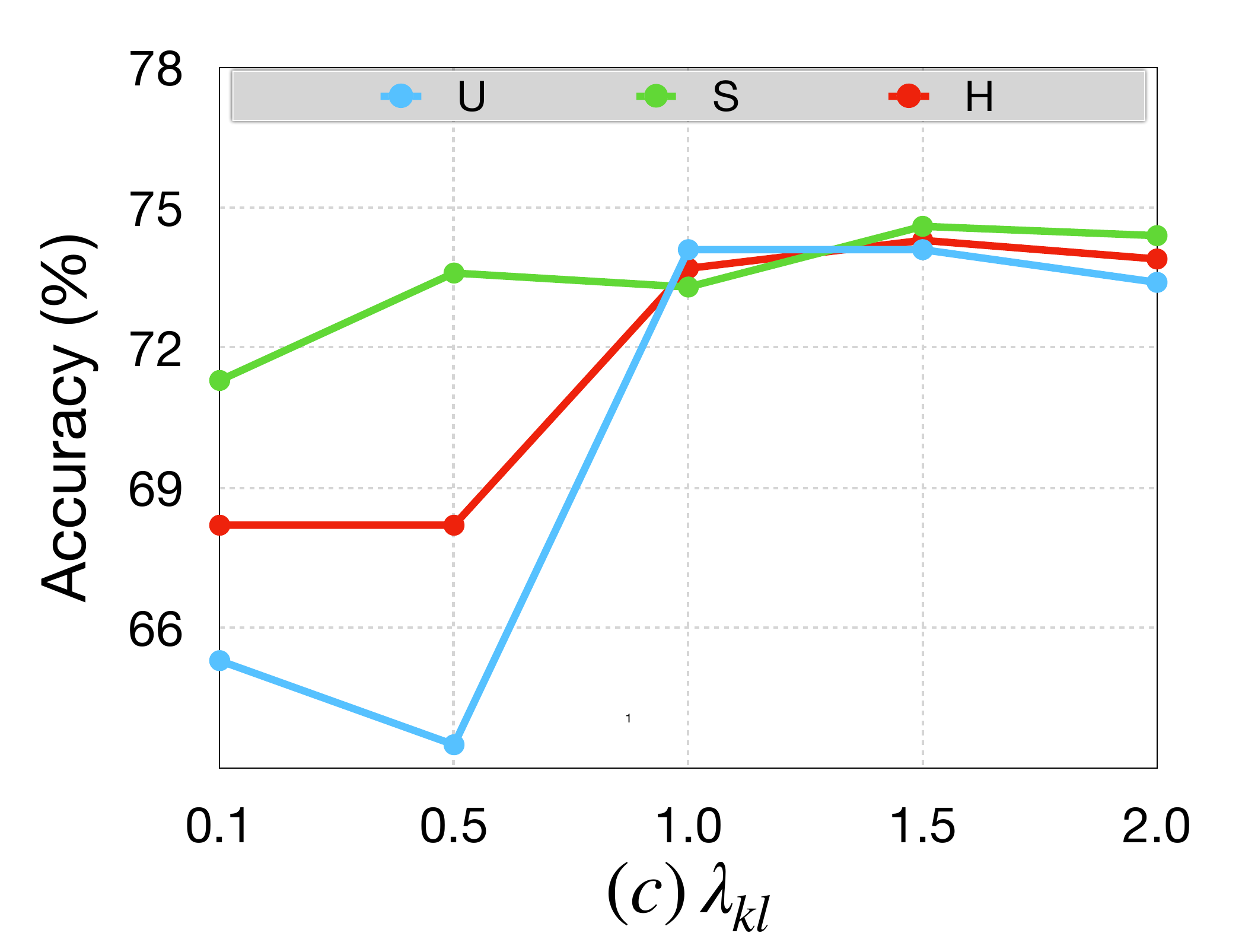}&
      \hspace{-5mm}\includegraphics[width=0.22\linewidth]{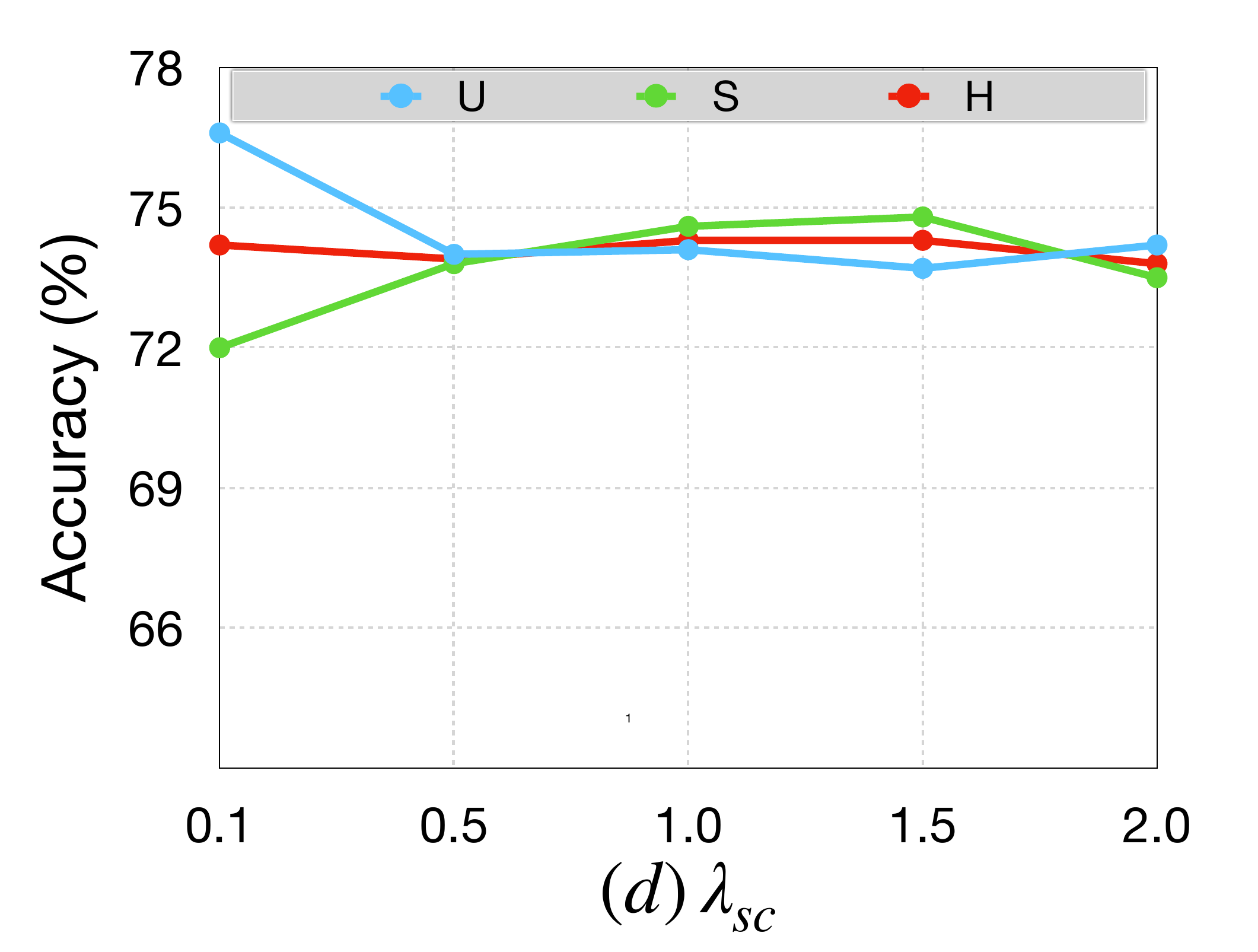}
    \end{tabular}\vspace{-4mm}
    \caption{Effect of (a) synthesized samples $N_{syn}$, (b) loss weights $\lambda_{con}$, (c) loss weights $\lambda_{kl}$, and (d) loss weights $\lambda_{sc}$ on CUB.}\label{fig:hyper-para}
  \end{center}
  \vspace{-5mm}
\end{figure*}

\subsection{Generalization Analysis of Visual Prior} 
\vspace{-2mm}
In our method, we learn a local bias and a fixed global prior vector representing domain visual prior knowledge to generalize to unseen classes. Therefore, we investigate the impact of different forms of prior knowledge on performance. The results are detailed in \cref{tab:ablation_prior}. ``Random" refers to the prior knowledge sampled from Gaussian distribution, ``VGSE" \cite{xu2022vgse} indicates that we take the semantic knowledge extracted from the visual representation as the prior, and ``Other domain" means using the global vector learned from SUN to transfer to CUB. The results indicate that our method utilizes the global prior and the local bias, which yields the best performance. Knowledge transfer from SUN to CUB suffers from a negative impact, underscoring the importance of dataset-specific global information. VGSE captures semantic side information from seen classes, limiting the performance of unseen classes.
\begin{table}[t]
  \small
  \centering
  \caption{We evaluate different forms of prior knowledge on CUB. The best result is marked in \bf{boldface}.}
      \begin{tabular}{l|c|c|ccc}
        \hline
        \multirow{2}*{Condition} &\multirow{2}*{Learnable} &\multicolumn{4}{c}{\bf{CUB}} \\
        \cline{3-6}
        &&\rm{Acc}&\rm{U} & \rm{S} & \rm{H}  \\
        \hline
        w/o prior &&84.9&69.1&75.5&72.2\\
        Random  & &85.6&72.7&74.7&73.4\\
        VGSE \cite{xu2022vgse} &\checkmark&85.5&69.2&72.3&70.7\\
        Other domain &\checkmark&84.0&66.7&72.0&69.3\\
        \bf{VADS (Ours)}  &\checkmark&\bf 86.8&74.1&74.6&\bf 74.3\\
        \hline
    \end{tabular}
      \label{tab:ablation_prior}
\end{table}

\subsection{Hyper-parameters Analysis}
We study the impact of different hyper-parameters of our VADS on the CUB dataset. \cref{fig:hyper-para}(a) shows the results of synthesizing different numbers per unseen class. The unseen class accuracy varies with the number of synthesized samples, and when $N_{syn} = $ 400, the performance is optimal. This result demonstrates that the features synthesized by our method alleviate the lack of unseen class data. Next, we evaluate the influence of individual loss weights (\textit{i.e.}, $\lambda_{con}$ of the $\mathcal L_{con}$, $\lambda_{kl}$ of the $\mathcal L_{kl}$ and $\lambda_{sc}$ of the $\mathcal L_{sc}$). The results are presented in \cref{fig:hyper-para}(b)(c)(d). The performance of $\bm S$ and $\bm U$ changes slightly as $\mathcal L_{con}$ increases. When $\mathcal L_{con}$ is set to 1.5, $\bm {H}$ obtains the maximum value. For $\mathcal L_{kl}$ loss, we find that larger $\lambda_{kl}$ achieves better performance because the prior distribution assumption of the data is crucial. Finally, $\mathcal L_{sc}$ loss forces the semantic update process to maintain semantic consistency. Its weight $\lambda_{sc}$ is insensitive to performance.

\subsection{Predefined Semantic Prototype vs Updated Semantic Prototype}
\begin{figure}[htbp]
  \centering
   \includegraphics[width=1.0\linewidth]{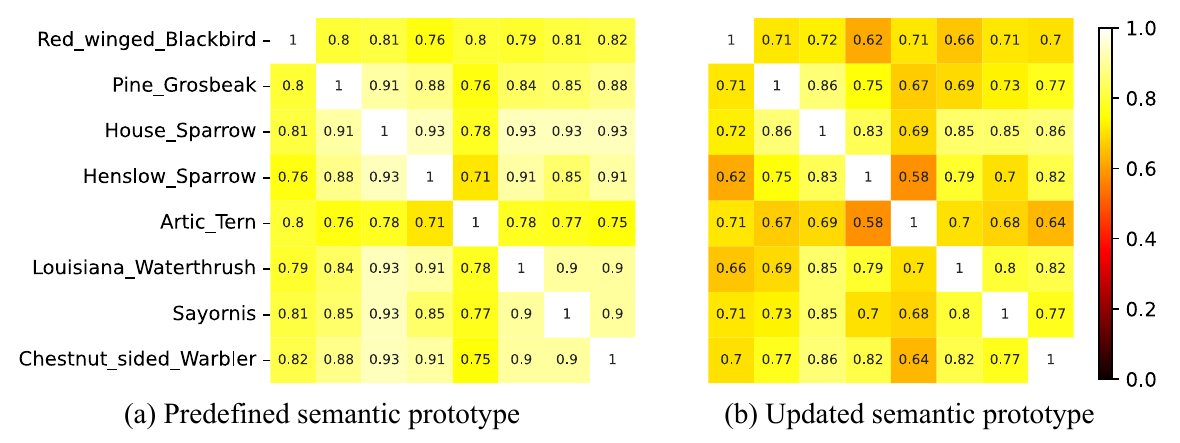}
   \caption{Visualization of the heatmap of semantic prototype similarity. We randomly select 10 classes on CUB.} 
   \label{fig:heatmap}
   \vspace{-5mm}
\end{figure}
To give a clearer insight into the predefined semantic prototypes and the semantic prototypes updated by our method, we randomly select 10 classes on CUB and calculate their cosine similarity. Then, we visualize them in \cref{fig:heatmap}. We observe that the similarity between the predefined semantic prototypes is very high. Our method dynamically refines the visual-semantic relationships of each instance based on visual information, making the updated semantic prototype easier to distinguish between categories and achieving more accurate semantic-visual alignment.

\section{Conclusion}
In this work, we propose a novel Visual-Augmented Dynamic Semantic prototype method (VADS) to boost the generator to synthesize reliable features of unseen classes. Considering that rich visual knowledge can effectively generalize to unseen classes, our proposed VADS fully leverages visual information. Specifically, we design a Visual-aware Domain Knowledge Learning module (VDKL) to acquire visual prior and a Vision-Oriented Semantic Updation module (VOSU) to dynamically update the predefined semantic prototype. Ultimately, we concatenate their output to form a dynamic semantic prototype, serving as the condition of the generator to learn accurate semantic-visual mapping and synthesize features of unseen classes. Extensive experiments demonstrate remarkable results in both CZSL and GZSL scenarios. In summary, our study provides a timely insight into reliable feature synthesis, improving the generalization to unseen classes. Additionally, tasks related to knowledge transfer can draw inspiration from this concept.

\section*{Acknowledgements}
This work is partially supported by National Key R\&D Program of China (2022YFC3301000).

{
    \small
    \bibliographystyle{ieeenat_fullname}
    \bibliography{main}
}

\end{document}